\documentclass[11pt]{article}
\usepackage[final]{acl}

\usepackage{times}
\usepackage{latexsym}
\usepackage[T1]{fontenc}
\usepackage[utf8]{inputenc}
\usepackage{microtype}
\usepackage{inconsolata}
\usepackage{graphicx}
\usepackage{booktabs}
\usepackage{float}
\usepackage{amsmath}
\usepackage{amssymb}
\usepackage{enumitem}
\usepackage{xspace}
\usepackage{tabularx}
\usepackage{array}
\newcolumntype{Y}{>{\raggedright\arraybackslash}X}
\setlist[itemize]{leftmargin=1.15em, itemsep=0.15em, topsep=0.25em}

\newcommand{\qwen}{\textsc{Qwen3-30B-A3B-Base}\xspace}
\newcommand{\mixtral}{\textsc{Mixtral-8x7B-v0.1}\xspace}
\newcommand{\cf}{\textsc{CounterFact}\xspace}

\newcommand{\moeout}{\mathrm{MoEOut}}
\newcommand{\clean}{\mathrm{clean}}
\newcommand{\noised}{\mathrm{noised}}
\newcommand{\patched}{\mathrm{patched}}
\newcommand{\rescue}{\mathrm{Rescue}}
\newcommand{\spec}{\mathrm{Spec}}
\newcommand{\edelta}{\delta_e}

\title{Expert-Aware Causal Tracing of Factual Recall\\
in Sparse MoE Language Models}

\author{
\textbf{Yuetian Lu}\textsuperscript{1,2,3},
\textbf{Ali Modarressi}\textsuperscript{1,3},
\textbf{Yihong Liu}\textsuperscript{1,3},
\textbf{Hinrich Sch{\"u}tze}\textsuperscript{1,3}
\\
\textsuperscript{1}Center for Information and Language Processing (CIS), LMU Munich \\
\textsuperscript{2}Ubiquitous Knowledge Processing Lab (UKP), Technical University of Darmstadt \\
\textsuperscript{3}Munich Center for Machine Learning (MCML)
\\
  \small{
    \textbf{Correspondence:} \href{mailto:yuetianlu@cis.lmu.de}{yuetianlu@cis.lmu.de}
}
}

\begin{document}
\maketitle
\begin{abstract}
Activation patching can identify a mixture-of-experts (MoE) block whose clean output restores a corrupted factual prediction.
However, because the block output combines contributions from multiple routed experts, block-level rescue does not establish whether the recovery localizes to an individual expert or depends on the routed expert set.
We study this question on single-token \cf contrasts by corrupting subject-token embeddings, restoring clean block outputs, and then restoring clean-minus-noised expert updates under fixed routing.
In \qwen, a discovery sweep selects layer 44, and held-out analysis identifies L44E069 as a recurrent routed contributor with positive specificity over same-layer active experts.
Its effect is fact-matched and improves true-token probability and rank, which explains part of the layer rescue. 
In \mixtral, the selected recurrent singleton is not specific; matched-size controls instead show specificity of the clean-routed top-2 set. 
These findings show that successful MoE-block restoration does not necessarily imply localization to a single expert.
\end{abstract}

\section{Introduction}

Factual recall is a common setting for causal analysis of language models: given a prompt such as \emph{The capital of France is}, the model should assign high probability to the correct object, \emph{Paris}. Causal tracing and activation patching study this behavior by corrupting a prediction and then restoring internal components from a clean run to test which components mediate the recovery \citep{meng2022locating,geva2023dissecting,vig2020causal,wang2023interpretability}. In dense transformers, these interventions often target residual-stream activations, attention outputs, or feed-forward modules.

Sparse mixture-of-experts (MoE) language models make this localization problem more fine-grained. A token is not processed by a single dense feed-forward module, but by a small routed subset of experts \citep{shazeer2017outrageously,fedus2022switch,jiang2024mixtral,yang2025qwen3}. Thus, a layer-level patch can show that an MoE block matters, but the patched block is an aggregate over routed expert updates. It does not answer which selected expert contributions support the factual prediction. This paper asks a simple question: \emph{When patching an MoE block restores a factual prediction, can part of that recovery be tied to a particular expert, or only to the routed set?}

We study this question in a controlled diagnostic setting using filtered single-token \cf \cite{meng2022locating} contrasts. For each fact, the input is a subject--relation prefix and the object is not included in the prompt. We measure factual preference as a true-vs-foil next-token logit difference, corrupt the run by adding noise to subject-token embeddings, and ask whether clean internal components restore the preference. Our protocol has two stages: first, we patch clean final-position MoE-block outputs into the noised run and sweep MoE layers on discovery cases; second, at the selected layer, we estimate routed expert contributions by ablation difference and patch an expert-level clean-minus-noised update into the noised run. We compare selected experts against other clean-active experts from the same layer.

The main result is asymmetric across two MoE backbones. On \qwen, the discovery sweep selects layer 44, and held-out expert-level tracing identifies L44E069 as a recurrent routed contributor with positive specificity over same-layer active controls. Additional held-out analyses show that its effect is fact-matched and improves true-token probability and rank, but is partial and nonunique: its mean rescue accounts for 51.4\% of mean L44 rescue. On \mixtral, the discovery sweep selects L19 within a broader mid-layer band, but L19E006 has negative singleton specificity. The clean-routed top-2 reconstructs the block-level effect and, unlike matched-size alternatives, exhibits routed-set specificity under fixed routing.

In summary, we apply expert-level activation patching to two MoE models and find that the same block-level rescue can have different expert-level structure: Qwen3 exhibits a partial recurrent singleton contribution, whereas Mixtral exhibits routed-set specificity without singleton specificity.

\section{Related Work}

\paragraph{Factual recall and causal tracing.}
Language models can express factual associations through cloze-style prompts \citep{petroni2019language}. Prior work links factual knowledge to feed-forward memories, knowledge neurons, and causal tracing interventions \citep{geva2021transformer,dai2021knowledge,meng2022locating,geva2023dissecting,wang-etal-2024-unveiling,liu-etal-2025-relation-specific}. 
Building on causal tracing and activation patching, we use forward-pass interventions to test whether clean internal components can restore a corrupted true-vs-foil factual preference.

\paragraph{Sparse MoE models and interpretability.}
Recent surveys organize mechanistic interpretability by the internal object being localized and the intervention used to test its role, while noting that sparse MoE routing requires architecture-specific treatment \citep{zhang-etal-2026-locate}. Sparse MoE language models route each token to a small subset of experts, increasing parameter count while keeping active computation sparse \citep{shazeer2017outrageously,lepikhin2020gshard,fedus2022switch,du2021glam,jiang2024mixtral,yang2025qwen3}. Recent work studies routing patterns, expert specialization, and MoE knowledge attribution \citep{fayyaz2026steering, bandarkar2025multilingualrouting,wang2026myth,li2025moeknowledgeattribution,gu2026moeedit}. Because the selected experts and routing weights are input-dependent, we distinguish MoE-block output patching from per-expert contribution patching. These studies ask different questions about routing, steering, editing, or attribution, so their scores are not directly comparable with our clean/corrupt restoration measure. We treat them as complementary and leave a common benchmark to future work.

\section{Method}

\subsection{\cf factual tracing setup}

Following the clean/corrupt intervention paradigm used in causal tracing and activation patching, we ask whether replacing a corrupted internal component with its clean counterpart restores the model's factual preference \citep{meng2022locating,vig2020causal,wang2023interpretability}. We use \cf facts introduced by \citet{meng2022locating}. Each record provides a subject, a relation-specific prompt template, a true object, and a counterfactual target object, which we use as the foil. For a prompt \(x\), true object \(a_t\), and foil object \(a_f\), we define:
\begin{equation}
    \Delta(x; a_t,a_f)
    =
    \ell_x(a_t)-\ell_x(a_f),
\end{equation}
where \(\ell_x(a)\) is the next-token logit for object token
\(a\). The object token is not included in the input; true
and foil objects only define the next-token contrast. For
example, if the input prompt prefix is \emph{``Virginia State Route 33 is located in''}; \emph{Virginia} is the true next-token object and \emph{Milan} is the foil.

We apply model-specific filters for single-token true/foil objects and sufficient clean/noise margins; Appendix~\ref{app:repro_details} gives the exact details.

\subsection{Subject-noise corrupted run}

To construct a corrupted run, we keep the text prompt unchanged but add Gaussian noise to the input embeddings of the subject-token span. If \(h_s\) is the embedding of a subject token, we use:
\begin{equation}
    h_s^{\noised}
    =
    h_s + \epsilon,
    \quad
    \epsilon \sim \mathcal{N}(0, \sigma^2 I),
\end{equation}
where \(\sigma\) is set to \(3.0\) times the embedding-matrix standard deviation. The resulting noised prompt has logit difference \(\Delta_{\noised}\). The subject-noise drop is:
\begin{equation}
    \Delta_{\clean}-\Delta_{\noised}.
\end{equation}

\subsection{MoE-block output patching}

For each clean prompt, we cache the final-position output of an MoE block at layer \(\ell\). This output is a hidden-state update vector produced by the routed MoE sublayer. We then run the noised prompt and replace the corresponding final-position MoE-block output with the clean one. The rescue score is:
\begin{equation}
    \rescue_\ell
    =
    \Delta_{\patched,\ell}
    -
    \Delta_{\noised}.
\end{equation}
We define rescue as a task-specific activation-restoration score: the change in the logit-difference metric after patching a clean component into a corrupted run. Positive rescue means that the clean MoE-block output restores part of the true-vs-foil factual preference.

We use a discovery/validation split. Layers are selected only on discovery cases and then evaluated as fixed hypotheses on validation cases.

\subsection{Expert-contribution patching}
\label{sec:expert_patching}

At a selected MoE layer, let \(\moeout(x)\) denote the final-position MoE-block output. For any expert \(e\), we define its routed update in a given run by the ablation difference:
\begin{equation}
    c_e(x)
    =
    \moeout(x) - \moeout_{\setminus e}(x),
\end{equation}
where \(\moeout_{\setminus e}\) is obtained by suppressing expert \(e\)'s contribution under the routing decision observed in that run. If \(e\) is not selected by the router in that run, then \(c_e(x)=0\). We do not reroute to a replacement expert nor renormalize the remaining expert weights. Accordingly, \(c_e(x)\), \(\edelta\), and their resulting rescue effects are \emph{conditional fixed-routing} quantities. They isolate the contribution of expert \(e\) as used under the observed routing decision, rather than the effect of removing \(e\) and allowing natural rerouting or renormalization. All expert-level interventions in this study are restricted to the final prediction position.

We define the expert patch vector
\begin{equation}
    \edelta
    =
    c_e(x_{\clean}) - c_e(x_{\noised}),
\end{equation}
and add \(\edelta\) to the noised MoE-block output at the final position. This asks whether replacing the noised routed expert update with its clean counterpart restores the true-vs-foil factual preference.

Expert selection must avoid two simple confounds. First, an expert can obtain high mean rescue by appearing in only one or two discovery cases. We therefore use recurrence-first selection: on discovery cases, we consider clean-active experts at the selected layer, require an expert to appear in at least a fixed number of discovery cases, and then select the recurrent candidate with the highest mean rescue. Second, a selected expert may not be special if other active experts in the same layer would rescue equally well. On validation cases, the selected expert is fixed and compared with same-source active-random experts, i.e., other clean-active experts selected by the router in the uncorrupted run for the same validation prompt and MoE layer.

We report selected-expert rescue and specificity:
\begin{equation}
    \spec
    =
    \rescue(e^\star)
    -
    \frac{1}{|\mathcal{R}|}
    \sum_{r\in\mathcal{R}}\rescue(r),
\end{equation}
where \(e^\star\) is the selected expert and \(\mathcal{R}\) is the active-random control set. Positive \(\spec\) means that the selected expert rescues more than other active same-layer experts for the same prompt.

\section{Experiments}

\subsection{Models}

We evaluate two sparse MoE language models. \qwen has 48 MoE layers, 128 experts per MoE layer, and top-8 routing \citep{yang2025qwen3}. \mixtral has 32 MoE layers, 8 experts per MoE layer, and top-2 routing \citep{jiang2024mixtral}.

\subsection{Data and splits}

We shuffle the full 21{,}919-record \cf pool with seed \(0\) and scan the shuffled order until collecting 256 usable cases per model, which we split into 128 discovery and 128 validation cases. Table~\ref{tab:filter_flow} reports the strict-filter collection flow. Because scanning stops once the quota is reached, the reported yield is a shuffled-prefix scan yield rather than a population-wide acceptance rate.

\begin{table}[t]
\centering
\small
\setlength{\tabcolsep}{5pt}
\begin{tabular}{lrrr}
\toprule
Model & Scanned prefix & Usable & Prefix yield \\
\midrule
Qwen3 & 390 & 256 & 65.6\% \\
Mixtral & 581 & 256 & 44.1\% \\
\bottomrule
\end{tabular}
\caption{Strict-filter collection flow from the shuffled \cf pool. Collection stops once 256 usable cases are obtained.}
\label{tab:filter_flow}
\end{table}

The single-token and subject-noise filters are model-specific because tokenizers and clean/noised logits differ across models. We therefore interpret the Qwen3--Mixtral contrast as two within-model case studies rather than as a controlled architectural comparison. The recurrence threshold in Section~\ref{sec:expert_patching} is set to half of the discovery split: a candidate expert must be clean-active in at least 64 of the 128 discovery cases. We additionally run a relaxed-filter scale-up with 512 usable cases per model, split into 256 discovery and 256 validation cases (Appendix~\ref{app:relaxed512}).

\subsection{Evaluation}

Layer rescue measures how much a clean MoE-block output restores the true-vs-foil contrast; expert rescue applies the same metric to one routed update. Specificity \(\spec\) compares the selected expert with same-source, same-layer active controls. Primary summaries include all held-out cases and zero-rescue rows; anchor-active analyses are labeled separately.

Unless noted otherwise, we report means, positive fractions, 5{,}000-resample case-bootstrap confidence intervals, and 10{,}000 two-sided sign-flip tests. Relation-clustered analyses first average within relation; the Mixtral matched-size controls use 10{,}000 bootstrap and 20{,}000 sign-flip samples. For Qwen3, we decompose true- and foil-logit changes and compare fact-matched with shuffled patches; for Mixtral, we compare routed sets with matched-size alternatives. Detailed definitions and sampling procedures are given in Appendices~\ref{app:repro_details}, \ref{app:qwen_outcome_controls}, and~\ref{app:mixtral_coalition}.

\section{Results}

\begin{figure*}[t]
\centering
\includegraphics[width=0.98\textwidth]{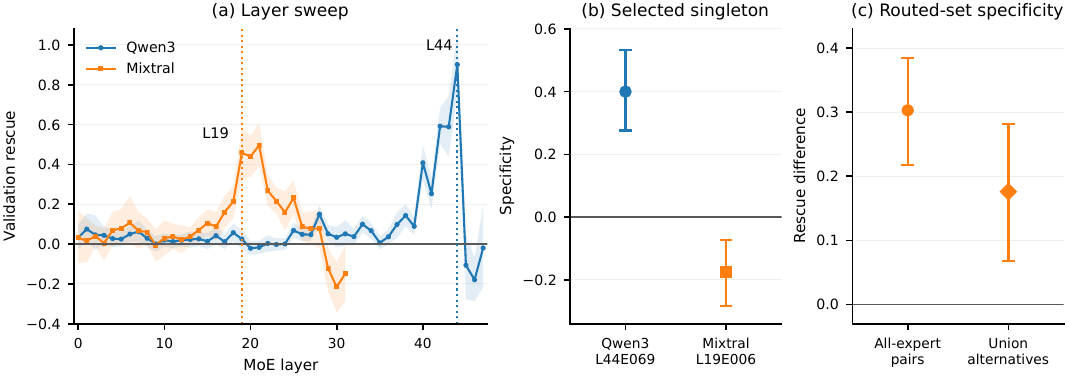}
\caption{Main validation pattern. Left: held-out MoE-block rescue across layers; Qwen3 L44 and Mixtral L19 were selected on discovery. Middle: selected-singleton specificity. Right: relation-clustered matched-size contrasts for Mixtral, comparing the clean-routed top-2 with all-expert and within-union alternatives. Error bars are 95\% CIs; positive values favor the actual routed set.}
\label{fig:all_layer_rescue_curves}
\end{figure*}

Figure~\ref{fig:all_layer_rescue_curves} summarizes the main held-out pattern: \textbf{Qwen3 exhibits a positive but partial recurrent singleton contribution, whereas Mixtral does not support singleton specificity and instead exhibits routed-set specificity under matched-size controls.}

\begin{table*}[t]
\centering
\small
\setlength{\tabcolsep}{3pt}
\begin{tabular*}{\textwidth}{@{\extracolsep{\fill}}lccccc@{}}
\toprule
Model & Layer & Layer rescue & Expert & Expert rescue & \(\spec\) \\
\midrule
Qwen3 & L44 & \(+0.901\) \([+0.752,+1.053]\) & L44E069 & \(+0.463\) \([+0.344,+0.590]\) & \(+0.400\) \([+0.276,+0.533]\) \\
Mixtral & L19 & \(+0.457\) \([+0.331,+0.579]\) & L19E006 & \(+0.099\) \([+0.018,+0.175]\) & \(-0.175\) \([-0.284,-0.072]\) \\
\bottomrule
\end{tabular*}
\caption{Main validation results on all 128 held-out cases per model. Expert rescue and \(\spec\) include zero-rescue rows; anchor-active-only analyses are labeled separately. \(\spec\) subtracts same-layer active-random rescue. Intervals are case-bootstrap 95\% CIs.}
\label{tab:main_results}
\end{table*}

\subsection{Layer-level tracing}
Table~\ref{tab:main_results} and the left panel of Figure~\ref{fig:all_layer_rescue_curves} summarize layer-level validation. The discovery sweep selects Qwen3 L44 and Mixtral L19, and both fixed sites have positive held-out rescue. Qwen3 L44 remains the strongest validation site. For Mixtral, L21 is slightly higher than the discovery-selected L19 on validation (\(+0.496\) versus \(+0.457\)); we therefore interpret L19 as a validated site within a broader mid-layer band rather than as a uniquely localized layer (Appendix~\ref{app:result_breakdowns}).

\subsection{Expert-level tracing}

Table~\ref{tab:main_results} and Figure~\ref{fig:all_layer_rescue_curves} show the expert-level results. On Qwen3, discovery selects L44E069, which has positive held-out specificity. Across cases with a valid gate-matched control, specificity is \(+0.459\); after equal-norm scaling it remains \(+0.188\). Thus, router weight and patch norm do not explain the advantage (Appendix~\ref{app:gate_norm_control}).

On Mixtral, L19E006 has negative specificity; over anchor-active cases, its equal-norm specificity is \(-0.062\). Routed-set patches reconstruct the L19 effect, and matched-size contrasts favor the actual routed set (Figure~\ref{fig:all_layer_rescue_curves}; Appendix~\ref{app:mixtral_coalition}). Mixtral therefore supports routed-set, but not selected-singleton, specificity.

\subsection{Additional held-out controls}
\label{sec:additional_controls}

All analyses in this subsection reuse the original validation cases with L44/L44E069 and L19/L19E006 fixed; no layer or expert is reselected.

\paragraph{Qwen3 effect content and fact matching.}
The mean L44E069 rescue of \(+0.463\) decomposes into a \(+0.394\) change in the true-token logit and a \(-0.069\) change in the foil-token logit. True-token probability and full-vocabulary rank improve in 67.2\% and 65.6\% of cases, respectively, although no case reaches top-1. A fact-matched expert patch rescues \(+0.463\), whereas a cross-fact shuffled patch rescues \(-0.032\); the relation-clustered matched-minus-shuffled contrast is \(+0.491\) (95\% CI \([+0.256,+0.753]\), \(p<.001\)). Thus, the intervention is fact-matched rather than a generic final-position restoration (Appendix~\ref{app:qwen_outcome_controls}).

\paragraph{Qwen3 is partial and nonunique.}
L44E069 accounts for 51.4\% of mean L44 rescue and has mean all-active rank 2.90 over 115 rankable validation cases. Replacing the expert-only patch with the full L44 patch adds \(+0.458\) relation-clustered rescue (95\% CI \([+0.309,+0.619]\), \(p<.001\)), leaving a substantial layer-level residual beyond the singleton. The mean rescue over all other clean-active experts is exactly the expected rescue under uniform random selection from that set; L44E069 exceeds this expectation by \(+0.441\) (95\% CI \([+0.309,+0.584]\)).

\paragraph{Mixtral routed-set specificity.}
The clean top-2 and routing-union effects are additive reconstruction checks; matched-size comparisons test whether the observed routed set is distinctive. Across all 128 validation cases, the clean top-2 exceeds the mean of eight sampled all-expert pairs by \(+0.303\) (95\% CI \([+0.217,+0.385]\), \(p<.001\)). On the 70 larger-union cases, it exceeds the mean of all within-union alternatives by \(+0.176\) (\(p=.0048\)). The union itself does not reliably outperform the clean top-2 (\(p=.929\)). Thus, the evidence supports additive sufficiency and average routed-set specificity (Appendix~\ref{app:mixtral_coalition}).

\paragraph{Relation-clustered uncertainty and scale.}
Relation-clustered inference preserves positive Qwen3 and negative Mixtral singleton specificity (\(p=.0012\) and \(p=.031\)). A relaxed 512-case analysis, whose validation set overlaps the strict set in only 15 cases, preserves the Qwen-positive/Mixtral-boundary pattern (Appendix~\ref{app:relaxed512}).

\section{Discussion}

\paragraph{What block-level rescue leaves unresolved.}
A successful block patch can hide two different patterns. In Qwen3, one recurrent expert makes a specific but partial contribution to the recovery. In Mixtral, the selected singleton is not specific, whereas the routed pair outperforms matched alternatives on average. This distinction appears only after intervening on expert updates; routing frequency alone does not establish it.

\paragraph{Interpreting the Qwen3--Mixtral asymmetry.}
The singleton result is counterintuitive under a simple dilution account: Qwen3 uses top-8 routing, yet one recurrent clean-active expert contribution has held-out specificity, whereas Mixtral uses top-2 routing but its selected recurrent singleton is not specific. The matched and equal-norm controls make a simple router-weight or patch-norm explanation less likely. Routing cardinality alone therefore does not determine the granularity at which a validated MoE-block effect resolves.

One architectural hypothesis is that Qwen3's finer expert partition (128 experts) may permit narrower recurrent roles, whereas Mixtral's coarser partition (8 experts) may distribute the relevant contribution across its routed set. Training choices, including load-balancing regularization and data curriculum, could also shape the observed granularity. These are hypotheses rather than identified causes: Qwen3 and Mixtral differ in architecture, tokenizer, training data, optimization, and retained cases, and our two-model design cannot isolate any one factor. This contrast concerns the presence or absence of a specific singleton contribution, not exclusive singleton mediation: a substantial Qwen3 layer-level residual remains beyond the L44E069 patch.

\section{Conclusion}

We used expert-level activation patching to ask how a rescued MoE-block signal is distributed across its routed experts. Under conditional fixed routing at the final prediction position, Qwen3 L44E069 is a recurrent routed contributor with held-out specificity, fact-matched effects, and improved true-token probability and rank, but it is partial and nonunique. Mixtral does not support selected-singleton specificity; instead, its clean-routed top-2 outperforms matched-size alternatives. The main empirical lesson is therefore that validated MoE-block restoration does not necessarily imply
localization to a single expert: the relevant contribution may include a specific recurrent singleton or resolve only at the routed-set level.

\section*{Ethical Considerations}

This work uses public model checkpoints and \cf-style factual diagnostics, and does not involve human subjects or private data. The method may support beneficial model debugging and factual-behavior analysis. Because \cf has limited relation and language coverage, erroneous localization or downstream editing may affect underrepresented factual associations unevenly. A potential dual-use risk is that more precise localization of factual behavior could support targeted manipulation, suppression, or editing of model outputs. Such uses should be evaluated carefully before deployment, especially for factual associations involving people, groups, languages, or regions. The submitted artifact contains diagnostic scripts and derived row-level results, but no model weights or deployment tool.

\section*{Limitations}

This study is a controlled factual-preference diagnostic rather than a full account of natural factual recall. The strict protocol uses 256 model-specific single-token true/foil \cf contrasts and one Gaussian draw per case for eligibility and intervention. The relaxed 512-case scale-up and fixed-hypothesis noise-scale analysis preserve the qualitative pattern, but we do not average over repeated within-case corruption draws. The outcome is defined against one designated foil. The additional analyses show true-logit, probability, and rank improvements, but no Qwen3 case reaches top-1, and we do not evaluate multiple foils, multi-token objects, or free generation.

Expert contributions are conditional fixed-routing effects at the final prediction position. They do not identify knowledge storage, natural rerouting behavior, or a complete subject-to-answer circuit. Our two-stage design decomposes the discovery-selected layer rather than exhaustively searching all layer--expert pairs, so useful mechanisms outside the selected site may be missed. In Mixtral, the selected L19 site belongs to a broader mid-layer band rather than a sharply unique layer.

Qwen3 and Mixtral differ in architecture, routing, tokenizer, training, and retained cases. With only two backbones, the observed contrast cannot be attributed causally to expert count, top-\(k\), load balancing, or training curriculum. We also do not include a dense-model baseline or a direct numerical benchmark against MoE attribution methods with different estimands. Isolating architectural causes would require a matched tokenization and training setup.

The relation and language coverage of \cf does not represent multilingual factual associations or many socially sensitive relation types. Erroneous localization or downstream editing may therefore affect underrepresented factual associations unevenly. Future work should evaluate broader factual datasets, repeated corruption draws, multi-token objects, position-wise interventions, more MoE backbones, and systematic coalition analyses. Reproducing the full experiments requires large sparse MoE checkpoints and implementation-specific hooks; the artifact provides scripts and row-level outputs, but the compute requirement remains nontrivial.

\section*{Acknowledgments}
\paragraph{Use of AI assistance.}
We used GPT-5.5 \citep{openai2026gpt55} as a coding and limited writing assistant. The authors checked the analyses, verified results, and are responsible for all content.

We thank Sascha Rothe from Google DeepMind for helpful discussions. We also
thank Google for its generous support through a
Google Cloud Credits grant. This work has been funded by the LOEWE Distinguished Chair ``Ubiquitous Knowledge Processing'', LOEWE initiative, Hesse, Germany (Grant Number: LOEWE/4a//519/05/00.002(0002)/81), and by the German Research Foundation (DFG) as part of grant GU 798/29-1 (UKP-SQuARE project) and grant SCHU 2246/14-1.

\bibliography{custom}

\appendix
\section{Reproducibility Details}
\label{app:repro_details}

This appendix lists protocol details and summary statistics for the 256-case experiments. Table~\ref{tab:repro_protocol_app} summarizes filtering, splitting, noise, and statistical settings. Table~\ref{tab:selected_expert_activity_app} reports selected-expert activity and zero-rescue rows, and Table~\ref{tab:relation_breakdown_app} summarizes retained relation counts. Exact model-specific \cf case IDs are provided in the supplementary artifact rather than typeset in the paper.

\begin{table}[h]
\centering
\scriptsize
\setlength{\tabcolsep}{3pt}
\begin{tabularx}{\columnwidth}{@{}lY@{}}
\toprule
Item & Setting \\
\midrule
Record order & \cf records shuffled with seed \(0\). \\
\addlinespace[0.15em]
Cases & 256 per model; 128 discovery and 128 validation. \\
\addlinespace[0.15em]
Filtering & Single-token true/foil objects; clean margin \(\geq 1.0\); subject-noise drop \(\geq 0.5\). \\
\addlinespace[0.15em]
Subject noise & One Gaussian draw per case; seed \(0+\texttt{case\_id}\); scale \(3.0\) times embedding-matrix std. \\
\addlinespace[0.15em]
Layer selection & MoE layers swept on discovery cases; selected layer fixed on validation cases. \\
\addlinespace[0.15em]
Expert selection & Candidate clean-active in at least 64 discovery cases; highest mean rescue selected. \\
\addlinespace[0.15em]
Active-random & Qwen3: 3 controls per case; Mixtral: 1 control per case. Controls are same-source, same-layer, clean-active experts excluding the selected expert. \\
\addlinespace[0.15em]
Statistics & Case-level: 5{,}000 bootstrap resamples and 10{,}000 two-sided sign-flip samples. Relation-clustered analyses use relation types as the resampling and sign-flip units. \\
\bottomrule
\end{tabularx}
\caption{Reproducibility protocol for the 256-case \cf experiments.}
\label{tab:repro_protocol_app}
\end{table}

\begin{table}[h]
\centering
\scriptsize
\setlength{\tabcolsep}{4pt}
\begin{tabular}{@{}lcccc@{}}
\toprule
Model & Expert & Disc. & Val. & Zero \\
\midrule
Qwen3 & L44E069 & 112/128 & 116/128 & 26 \\
Mixtral & L19E006 & 91/128 & 83/128 & 54 \\
\bottomrule
\end{tabular}
\caption{Selected-expert activity and zero-rescue rows in expert-level validation. Disc. and Val. count discovery and validation cases in which the selected expert has a nonzero clean contribution under the observed routing decision. Zero counts cases with exactly zero selected-expert rescue; this need not equal the number of clean-inactive cases because clean and noised routing can differ. Zero-rescue rows are retained in the all-case expert rescue and specificity summaries.}
\label{tab:selected_expert_activity_app}
\end{table}

Clean activity and zero rescue are distinct quantities. If the selected expert is inactive in both the clean and noised runs, its patch vector is zero. If it is inactive only in the clean run but active in the noised run, then \(\edelta=-c_e(x_{\noised})\), and the intervention can be nonzero. We retain all validation cases in the primary expert-level summaries; anchor-active subsets are labeled separately in the corresponding controls.

\begin{table}[h]
\centering
\scriptsize
\setlength{\tabcolsep}{3pt}
\begin{tabularx}{\columnwidth}{@{}lY@{}}
\toprule
Model & Most frequent retained relations \\
\midrule
Qwen3 & P17:21, P103:16, P136:14, P178:13, P27:13, P131:12, P159:12, P30:12, P20:11, P176:11, P740:9, P364:9 \\
Mixtral & P17:28, P103:26, P27:16, P178:15, P495:15, P136:15, P740:15, P106:14, P1412:14, P131:11, P20:10, P364:9 \\
\bottomrule
\end{tabularx}
\caption{Most frequent retained \cf relations after model-specific filtering. Counts are over the 256 usable cases for each model.}
\label{tab:relation_breakdown_app}
\end{table}

\section{Result Breakdowns}
\label{app:result_breakdowns}

This appendix provides additional validation summaries that support the main results. Table~\ref{tab:relation_effects_app} reports relation-wise expert-level effects for frequent retained relations. Table~\ref{tab:layer_sharpness_app} quantifies the layer-sweep sharpness, and Table~\ref{tab:expert_layer_fraction_app} relates selected-expert rescue to selected-layer MoE-block rescue.

\begin{table*}[t]
\centering
\scriptsize
\setlength{\tabcolsep}{4pt}
\begin{tabular}{llrrrr}
\toprule
Model & Relation & \(n\) & Expert rescue & Specificity & Pos. frac. \\
\midrule
Qwen3 & P103 native lang. & 12 & \(+0.427\) & \(+0.406\) & 0.83 \\
Qwen3 & P27 citizenship & 9 & \(+0.826\) & \(+0.780\) & 1.00 \\
Qwen3 & P136 genre & 7 & \(+0.018\) & \(-0.000\) & 0.57 \\
Qwen3 & P131 admin. location & 6 & \(+1.240\) & \(+1.160\) & 0.83 \\
Qwen3 & P17 country & 6 & \(+1.073\) & \(+0.910\) & 0.67 \\
Qwen3 & P30 continent & 6 & \(+0.292\) & \(+0.312\) & 0.50 \\
Qwen3 & P1412 language used & 6 & \(+0.198\) & \(+0.163\) & 0.67 \\
Qwen3 & P413 position & 6 & \(+0.042\) & \(-0.174\) & 0.17 \\
Qwen3 & P937 work location & 5 & \(+1.387\) & \(+1.350\) & 1.00 \\
Qwen3 & P159 headquarters & 5 & \(+1.012\) & \(+0.988\) & 0.80 \\
\midrule
Mixtral & P103 native lang. & 12 & \(+0.312\) & \(+0.206\) & 0.67 \\
Mixtral & P17 country & 11 & \(+0.187\) & \(-0.272\) & 0.09 \\
Mixtral & P1412 language used & 9 & \(+0.031\) & \(-0.076\) & 0.33 \\
Mixtral & P27 citizenship & 9 & \(+0.049\) & \(-0.132\) & 0.22 \\
Mixtral & P740 formation loc. & 9 & \(+0.146\) & \(-0.167\) & 0.56 \\
Mixtral & P495 origin country & 9 & \(+0.094\) & \(-0.358\) & 0.11 \\
Mixtral & P178 developer & 9 & \(+0.000\) & \(-0.368\) & 0.00 \\
Mixtral & P136 genre & 9 & \(+0.111\) & \(-0.562\) & 0.11 \\
Mixtral & P106 occupation & 7 & \(+0.223\) & \(+0.179\) & 0.71 \\
Mixtral & P131 admin. location & 6 & \(-0.016\) & \(-0.557\) & 0.00 \\
\bottomrule
\end{tabular}
\caption{Relation-wise expert-level validation summary for frequent retained relations. Expert means include all held-out rows assigned to each relation, including zero-rescue rows. Specificity is selected-expert rescue minus active-random rescue. Relation counts are model-specific after filtering.}
\label{tab:relation_effects_app}
\end{table*}

\begin{table}[h]
\centering
\scriptsize
\setlength{\tabcolsep}{4pt}
\begin{tabular}{lrrrr}
\toprule
Model & Top layer & Top rescue & Next layer & Gap \\
\midrule
Qwen3 & L44 & \(+0.901\) & L42 \(+0.592\) & \(+0.309\) \\
Mixtral & L21 & \(+0.496\) & L19 \(+0.457\) & \(+0.038\) \\\bottomrule
\end{tabular}
\caption{Layer-sweep sharpness on validation cases. Gap is the difference between the top validation rescue and the next-highest layer. Mixtral's smaller gap reflects a broader mid-layer band.}
\label{tab:layer_sharpness_app}
\end{table}

\begin{table}[h]
\centering
\scriptsize
\setlength{\tabcolsep}{4pt}
\begin{tabular}{llr}
\toprule
Model & Quantity & Value \\
\midrule
Qwen3 & Layer rescue & \(+0.901\) \\
Qwen3 & Expert/layer & \(+0.515\) \([+0.411,+0.613]\) \\
Qwen3 & \(\spec\)/layer & \(+0.444\) \([+0.323,+0.560]\) \\
\midrule
Mixtral & Layer rescue & \(+0.457\) \\
Mixtral & Expert/layer & \(+0.216\) \([+0.044,+0.388]\) \\
Mixtral & \(\spec\)/layer & \(-0.383\) \([-0.645,-0.159]\) \\
\bottomrule
\end{tabular}
\caption{Expert-level rescue relative to selected-layer MoE-block rescue. Ratios are computed as ratios of validation-case means, with bootstrap confidence intervals. \(\spec\)/layer compares specificity \(\spec\) to selected-layer rescue.}
\label{tab:expert_layer_fraction_app}
\end{table}

\section{Matched and Equal-Norm Expert Controls}
\label{app:gate_norm_control}

\subsection{Qwen3 Gate-Weight-Matched and Equal-Norm Control}
To test whether the Qwen3 L44E069 result is driven by router weight or patch-vector magnitude, we run a stricter validation-only control. Let \(e^\star\) be L44E069 and let \(A(x)\) be the set of other clean-active layer-44 experts for the same source prompt. For each validation case where \(e^\star\) is clean-active, we choose the gate-weight-matched control
\begin{equation}
    r^\star
    =
    \arg\min_{r\in A(x)}
    \left|w_{r}(x_{\clean}) - w_{e^\star}(x_{\clean})\right|.
\end{equation}
The raw gate-matched specificity is
\begin{equation}
    \spec_{\mathrm{gate}}
    =
    \rescue(e^\star) - \rescue(r^\star).
\end{equation}
For the equal-norm version, we scale the selected and control patch vectors to the smaller original norm:
\begin{equation}
    \tilde{\delta}_e
    =
    \delta_e
    \frac{
        \min(\|\delta_{e^\star}\|,\|\delta_{r^\star}\|)
    }{
        \|\delta_e\|
    },
\end{equation}
and compute the same specificity contrast after patching \(\tilde{\delta}_{e^\star}\) and \(\tilde{\delta}_{r^\star}\). This control is evaluated on 116 of 128 Qwen validation cases where L44E069 is clean-active and a valid same-layer active control exists. Table~\ref{tab:gate_norm_control} reports the results.

\begin{table*}[t]
\centering
\scriptsize
\setlength{\tabcolsep}{5pt}
\begin{tabular}{lrr}
\toprule
Quantity & Raw & Equal-norm \\
\midrule
Selected rescue & \(+0.513\) \([+0.383,+0.651]\) & \(+0.239\) \([+0.177,+0.302]\) \\
Matched control & \(+0.054\) \([+0.017,+0.091]\) & \(+0.051\) \([+0.016,+0.086]\) \\
Specificity & \(+0.459\) \([+0.322,+0.607]\) & \(+0.188\) \([+0.119,+0.259]\) \\
\bottomrule
\end{tabular}
\caption{Qwen3 L44E069 gate-weight-matched and equal-norm control on 116 validation cases. The matched control is the clean-active layer-44 expert with closest clean router weight. Equal-norm rows scale both patch vectors to the smaller original norm before patching.}
\label{tab:gate_norm_control}
\end{table*}

For the all-active comparison, let \(A(x)\) be the clean-active layer-44 expert set and \(e^\star=\mathrm{L44E069}\). We define
\begin{align}
A^-(x) &= A(x)\setminus\{e^\star\},\\
\bar{R}_{A^-}(x)
&=
\frac{1}{|A^-(x)|}
\sum_{r\in A^-(x)}\rescue(r),\\
\spec_{\mathrm{all}}(e^\star)
&=
\rescue(e^\star)-\bar{R}_{A^-}(x).
\end{align}
Because \(\bar{R}_{A^-}(x)\) averages over all alternative clean-active experts, it is exactly the expected rescue of an expert sampled uniformly from \(A^-(x)\). Thus, \(\spec_{\mathrm{all}}\) directly compares L44E069 with uniform random clean-active selection. Table~\ref{tab:qwen_all_active_percentile} additionally compares L44E069 against all clean-active layer-44 experts in each validation case where an all-active rank comparison is available.

\begin{table}[h]
\centering
\scriptsize
\setlength{\tabcolsep}{4pt}
\begin{tabular}{lr}
\toprule
Metric & Value \\
\midrule
Ranked validation cases & 115 \\
Top-1 among active experts & 53 / 115 \\
Top-2 among active experts & 74 / 115 \\
Mean rank & \(2.90\) \\
Mean percentile & \(0.73\) \\
Selected minus all-other active & \(+0.441\) \([+0.309,+0.584]\) \\
\bottomrule
\end{tabular}
\caption{Qwen3 L44E069 rescue rank among all clean-active layer-44 experts on 115 rankable validation cases. ``Selected minus all-other active'' is the advantage over the uniform-random expectation on the other clean-active experts.}
\label{tab:qwen_all_active_percentile}
\end{table}

\subsection{Mixtral Top-2 Active-Pair Equal-Norm Check}

For Mixtral, top-2 routing makes the active-random control the unique other clean-active expert whenever the selected expert is clean-active. Therefore gate-weight matching is degenerate: the gate-matched control and active-random control are the same expert. We instead use an equal-norm active-pair check, scaling the L19E006 patch vector and the other active expert's patch vector to the smaller original norm before patching. Table~\ref{tab:mixtral_equalnorm_active_pair} reports this check.

\begin{table}[h]
\centering
\scriptsize
\setlength{\tabcolsep}{4pt}
\begin{tabular}{lrrr}
\toprule
Metric & \(n\) & Mean & 95\% CI \\
\midrule
Raw active-pair specificity & 83 & \(-0.081\) & \([-0.214,+0.042]\) \\
Selected equal-norm rescue & 83 & \(+0.046\) & \([-0.007,+0.096]\) \\
Other active expert equal-norm rescue & 83 & \(+0.108\) & \([+0.047,+0.172]\) \\
Equal-norm active-pair specificity & 83 & \(-0.062\) & \([-0.130,-0.003]\) \\
\bottomrule
\end{tabular}
\caption{Mixtral L19E006 active-pair equal-norm check on anchor-active validation cases. Specificity is L19E006 rescue minus the rescue of the unique other clean-active layer-19 expert. The selected expert remains below the other active expert after equal-norm scaling.}
\label{tab:mixtral_equalnorm_active_pair}
\end{table}

\section{Qwen3 Outcome and Cross-Fact Controls}
\label{app:qwen_outcome_controls}

All analyses in this section use the original 128 Qwen3 validation cases with L44 and L44E069 fixed. We decompose the change induced by the expert patch into true- and foil-token logit changes, record whether true-token probability and full-vocabulary rank improve, and test whether the true token becomes top-1. We additionally compare each fact-matched L44E069 patch vector with a cross-fact shuffled patch vector under the fixed permutation reported in the artifact.

\begin{table}[t]
\centering
\scriptsize
\setlength{\tabcolsep}{4pt}
\begin{tabularx}{\columnwidth}{@{}Yr@{}}
\toprule
Metric & Value \\
\midrule
True-token logit change & \(+0.394\) \\
Foil-token logit change & \(-0.069\) \\
True-token probability improved & 67.2\% \\
Full-vocabulary rank improved & 65.6\% \\
True token became top-1 & 0\% \\
Fact-matched expert rescue & \(+0.463\) \\
Cross-fact shuffled rescue & \(-0.032\) \\
Matched minus shuffled & \(+0.491\) \([+0.256,+0.753]\), \(p<.001\) \\
Expert/layer mean-rescue ratio & 51.4\% \\
Full L44 minus expert-only patch & \(+0.458\) \([+0.309,+0.619]\), \(p<.001\) \\
\bottomrule
\end{tabularx}
\caption{Qwen3 held-out outcome decomposition, cross-fact control, and residual layer effect. Confidence intervals and \(p\)-values for matched-minus-shuffled and full-layer-minus-expert contrasts are relation-clustered.}
\label{tab:qwen_outcome_controls}
\end{table}

The true-vs-foil rescue is therefore driven primarily by increasing the true-token logit rather than by suppressing only the designated foil. The probability, rank, and shuffled-delta results show that the effect is fact-matched but does not change greedy top-1 output. The residual full-layer effect confirms that L44E069 is a partial and nonunique contributor.

\section{Qwen3 Expert-Selection Stability}
\label{app:expert_stability}

We test whether the Qwen3 selected expert is sensitive to the discovery/validation split or recurrence threshold. Reusing the 256 usable Qwen3 cases, we vary the split seed over \(\{0,1,2,3,4\}\) and the recurrence threshold over \(\{32,48,64,80,96\}\). Across all 25 settings, L44E069 is selected every time. Averaged over split seeds and thresholds, validation rescue is \(+0.398\) and specificity is \(+0.344\).

\paragraph{Relation-held-out selection.}
We also test whether L44E069 is selected when entire relation types are held out from expert selection. We split retained Qwen3 relations into five folds, select the recurrent layer-44 expert using only non-held-out relations, and evaluate on the held-out relations. L44E069 is selected in all five folds. Table~\ref{tab:qwen_relation_heldout} reports the held-out rescue and specificity.

\section{Noise-Scale Sensitivity}
\label{app:noise_sensitivity}

We run a fixed-hypothesis sensitivity check for the Qwen3 result by keeping the selected layer and expert fixed to L44 and L44E069, and varying only the subject-noise scale \(\sigma\). This check does not re-select layers or experts; it asks whether the fixed L44E069 intervention remains positive under different corruption strengths.

Table~\ref{tab:qwen_noise_scale_sensitivity} summarizes the fixed-hypothesis results.

\section{Relaxed-Filter Scale-Up}
\label{app:relaxed512}

We run an additional relaxed-filter scale-up to test whether the main pattern depends on the stricter 256-case filtering regime. The clean-margin threshold is lowered from \(1.0\) to \(0.5\), and the subject-noise-drop threshold is lowered from \(0.5\) to \(0.25\), yielding 512 usable cases per model. We use the same discovery/validation protocol and the same half-discovery recurrence threshold.

Only 15 of 256 relaxed validation cases overlap with the main strict set. Table~\ref{tab:relaxed512_new_cases_app} shows that the relaxed-new validation cases preserve the same Qwen-positive/Mixtral-boundary pattern.

\section{Mixtral Routed-Set Reconstruction and Matched-Size Controls}
\label{app:mixtral_coalition}

We first test whether additive clean-minus-noised contributions from the observed routed experts reconstruct the positive L19 MoE-block effect. We then ask the more discriminating question of whether the actual routed set outperforms matched-size alternatives. All analyses use the original 128 validation cases with L19 fixed.

Let \(S_c(x)\) be the experts selected by the router in the clean run at layer 19, and let \(S_n(x)\) be the experts selected at the same position in the noised run. Recall that \(\edelta\) denotes expert \(e\)'s clean-minus-noised patch vector. We define the coalition patches as

\begin{align}
    D_{\mathrm{top2}}(x)
    &=
    \sum_{e\in S_c(x)} \edelta,\\
    D_{\mathrm{union}}(x)
    &=
    \sum_{e\in S_c(x)\cup S_n(x)} \edelta.
\end{align}
The first patch restores the clean-routed expert coalition; the second also accounts for experts that enter or leave the routed set under subject noise. Table~\ref{tab:mixtral_coalition} reports the resulting coalition-patching rescue values.

\paragraph{Matched-size baselines.}
All random baselines use base seed \(0\). For each case, a NumPy generator is initialized with seed \(100000+\texttt{case\_id}\).

For the all-expert pair baseline, we enumerate the \(\binom{8}{2}=28\) unordered pairs over the eight layer-19 experts, exclude the clean-routed pair, and sample eight of the remaining 27 pairs uniformly without replacement. The per-case baseline is the mean rescue over these eight sampled pairs.

For the within-union comparison, no random subsampling is used. Whenever \(|S_c(x)\cup S_n(x)|>2\), we enumerate every size-2 pair from the clean/noised routing union, exclude the clean-routed pair, and average rescue over all remaining alternatives. The 70 applicable cases comprise 61 routing unions of size three, which yield two alternatives each, and nine routing unions of size four, which yield five alternatives each.

For the routing-union baseline, we enumerate all subsets of size \(|S_c(x)\cup S_n(x)|\) over the eight layer-19 experts, exclude the actual routing union, and sample eight same-cardinality coalitions uniformly without replacement.

For relation-clustered inference, case-level contrasts are first averaged within relation. Confidence intervals use 10{,}000 bootstrap resamples and \(p\)-values use 20{,}000 two-sided sign-flip samples, both with seed \(0\).

The clean-routed top-2 and routing union outperform their respective matched-size baselines, supporting routed-set specificity relative to matched-size alternatives and additive sufficiency. The within-union result compares the clean-routed pair with the \emph{mean} rescue over all alternatives; it does not imply that the clean pair outperforms every individual alternative. These results do not establish non-additive synergy, exhaustive coalition attribution, natural rerouting behavior, or a complete factual-recall circuit.

\clearpage

\makeatletter
\setlength{\@dblfptop}{0pt}
\setlength{\@dblfpsep}{8pt}
\setlength{\@dblfpbot}{0pt plus 1fil}
\makeatother

\begingroup
\setlength{\abovecaptionskip}{3pt}
\setlength{\belowcaptionskip}{3pt}
\setlength{\textfloatsep}{6pt}
\setlength{\floatsep}{6pt}

\begin{table*}[!t]
\centering
\vspace*{-0.5em}

\begin{minipage}[t]{0.40\textwidth}
\centering
\scriptsize
\setlength{\tabcolsep}{4pt}
\begin{tabular}{lr}
\toprule
Metric & Value \\
\midrule
Folds & 5 \\
Selected & 5 / 5 \\
Held-out cases & 256 \\
Active cases & 229 / 256 \\
Rescue & \(+0.443\) \([+0.343,+0.545]\) \\
Specificity & \(+0.388\) \([+0.297,+0.484]\) \\
\bottomrule
\end{tabular}
\caption{Qwen3 relation-held-out expert-selection check over five relation folds.}
\label{tab:qwen_relation_heldout}
\end{minipage}
\hfill
\begin{minipage}[t]{0.56\textwidth}
\centering
\scriptsize
\setlength{\tabcolsep}{4pt}
\begin{tabular}{lrrrr}
\toprule
\(\sigma\) & Active & Drop & Rescue & Pos. \\
\midrule
1.0 & 115/128 & \(+1.259\) & \(+0.098\) \([+0.040,+0.165]\) & 0.43 \\
2.0 & 115/128 & \(+4.909\) & \(+0.406\) \([+0.280,+0.546]\) & 0.55 \\
3.0 & 115/128 & \(+5.697\) & \(+0.459\) \([+0.339,+0.587]\) & 0.66 \\
4.0 & 115/128 & \(+6.004\) & \(+0.478\) \([+0.356,+0.605]\) & 0.67 \\
\bottomrule
\end{tabular}
\caption{Qwen3 fixed-hypothesis noise-scale sensitivity for L44E069 on all 128 validation cases. Active counts cases where L44E069 has a nonzero clean contribution in the replayed run. Drop is mean clean-minus-noised logit-difference drop; Rescue is mean L44E069 rescue with 95\% bootstrap confidence intervals; Pos. is the fraction of all validation cases with positive rescue.}
\label{tab:qwen_noise_scale_sensitivity}
\end{minipage}

\vspace{0.75em}

\begin{minipage}{0.96\textwidth}
\centering
\scriptsize
\setlength{\tabcolsep}{4pt}
\begin{tabular*}{0.96\textwidth}{@{\extracolsep{\fill}}lccccc@{}}
\toprule
Model & Layer rescue & Expert & Disc. active & Expert rescue & \(\spec\) \\
\midrule
Qwen3 & \(+0.846\) \([+0.760,+0.934]\) & L44E069 & 230/256 & \(+0.454\) \([+0.367,+0.545]\) & \(+0.413\) \([+0.318,+0.511]\) \\
Mixtral & \(+0.421\) \([+0.352,+0.489]\) & L19E006 & 174/256 & \(+0.114\) \([+0.041,+0.187]\) & \(-0.155\) \([-0.247,-0.067]\) \\
\bottomrule
\end{tabular*}
\caption{Relaxed-filter discovery results on 256 discovery cases per model, used to select the layer and recurrent expert. The clean-margin threshold is \(0.5\), the subject-noise-drop threshold is \(0.25\), and the recurrence threshold is half of the discovery split. Held-out validation results are reported in Table~\ref{tab:relaxed512_new_cases_app}.}
\label{tab:relaxed512_app}
\end{minipage}

\vspace{0.75em}

\begin{minipage}{0.96\textwidth}
\centering
\tiny
\setlength{\tabcolsep}{3pt}
\begin{tabular*}{0.96\textwidth}{@{\extracolsep{\fill}}llrrrrr@{}}
\toprule
Model & Subset & \(n_L\) & Layer & \(n_E\) & Expert & \(\spec\) \\
\midrule
Qwen3 & All relaxed val. & 256 & \(+0.833\) \([+0.714,+0.962]\) & 137 & \(+0.400\) \([+0.293,+0.516]\) & \(+0.360\) \([+0.239,+0.487]\) \\
Qwen3 & Relaxed-new val. & 241 & \(+0.860\) \([+0.739,+0.989]\) & 130 & \(+0.420\) \([+0.310,+0.541]\) & \(+0.380\) \([+0.259,+0.509]\) \\
Mixtral & All relaxed val. & 256 & \(+0.454\) \([+0.360,+0.557]\) & 137 & \(+0.119\) \([-0.003,+0.240]\) & \(-0.124\) \([-0.270,+0.014]\) \\
Mixtral & Relaxed-new val. & 241 & \(+0.459\) \([+0.360,+0.561]\) & 130 & \(+0.123\) \([-0.010,+0.250]\) & \(-0.128\) \([-0.280,+0.014]\) \\
\bottomrule
\end{tabular*}
\caption{Relaxed-filter held-out validation results, split by overlap with the main strict set. ``Relaxed-new'' cases occur in the relaxed-filter validation split but not in the strict 256-case set. \(n_L\) is the number of layer-level cases; \(n_E\) is the number of expert-level contrast rows over which the corresponding expert rescue and specificity means are computed.}
\label{tab:relaxed512_new_cases_app}
\end{minipage}

\vspace{0.75em}

\begin{minipage}{0.96\textwidth}
\centering
\scriptsize
\setlength{\tabcolsep}{5pt}
\begin{tabular}{lrr}
\toprule
Patch & Rescue & Pos. frac. \\
\midrule
Clean top-2 coalition
& \(+0.461\) \([+0.343,+0.572]\)
& 0.75 \\
Routing-union coalition
& \(+0.490\) \([+0.367,+0.613]\)
& 0.75 \\
\bottomrule
\end{tabular}
\caption{Mixtral L19 additive routed-set reconstruction on validation cases. The clean top-2 sums patch vectors for the clean-routed experts; the routing union sums patch vectors over the union of clean- and noised-routed experts. Because these patches follow the additive MoE decomposition, this table is interpreted as a reconstruction check rather than evidence of routed-set specificity.}
\label{tab:mixtral_coalition}
\end{minipage}

\end{table*}
\endgroup

\begin{table*}[!t]
\centering
\scriptsize
\setlength{\tabcolsep}{4pt}
\begin{tabularx}{\textwidth}{@{}Yrrrrr@{}}
\toprule
Contrast & Cases & Relations & Mean & 95\% CI & \(p\) \\
\midrule
Clean top-2 -- mean of 8 all-expert pairs
& 128
& 26
& \(+0.303\)
& \([+0.217,+0.385]\)
& \(<.001\) \\
Clean top-2 -- mean of all union alternatives
& 70
& 23
& \(+0.176\)
& \([+0.068,+0.281]\)
& \(.0048\) \\
Routing union -- mean of 8 same-size coalitions
& 128
& 26
& \(+0.275\)
& \([+0.188,+0.360]\)
& \(<.001\) \\
Routing union -- clean top-2
& 128
& 26
& \(+0.002\)
& \([-0.043,+0.041]\)
& \(.929\) \\
\bottomrule
\end{tabularx}
\caption{Mixtral matched-size routed-set controls. Each contrast is first computed per case against the corresponding sampled or exhaustively enumerated baseline. Means weight relations equally after averaging case-level contrasts within relation. Confidence intervals use 10{,}000 bootstrap resamples and \(p\)-values use 20{,}000 two-sided sign-flip samples over relation means, with seed \(0\).}
\label{tab:mixtral_matched_size}
\end{table*}

\end{document}